\documentclass[10pt,twocolumn,letterpaper]{article}

\usepackage{iccv}              % To produce the CAMERA-READY version

\usepackage{graphicx}
\usepackage{caption}
\usepackage{lipsum}
\usepackage{multirow}
\usepackage{array}
\usepackage{wrapfig}
\usepackage{makecell}

\DeclareMathOperator*{\argmin}{arg\,min}

\definecolor{iccvblue}{rgb}{0.21,0.49,0.74}
\usepackage[pagebackref,breaklinks,colorlinks,allcolors=iccvblue]{hyperref}

\title{EvAnimate: Event-conditioned Image-to-Video Generation for Human Animation}

\author{
Qiang Qu\textsuperscript{1} \quad
Ming Li\textsuperscript{2} \quad
Xiaoming Chen\textsuperscript{2} \quad
Tongliang Liu\textsuperscript{1} \quad \\
\textsuperscript{1}University of Sydney \quad
\textsuperscript{2} Beijing Technology and Business University\\
{\tt\small vincent.qu.cs@gmail.com},\; {\tt\small 15190293355@163.com},\\
{\tt\small xiaoming.chen@btbu.edu.cn},\; {\tt\small tongliang.liu@sydney.edu.au}
}
\begin{document}

\twocolumn[{%
\renewcommand\twocolumn[1][]{#1}%
\maketitle
\begin{center}
\vspace{-2em}
    \url{https://potentialming.github.io/projects/EvAnimate}
    \includegraphics[width=\linewidth]{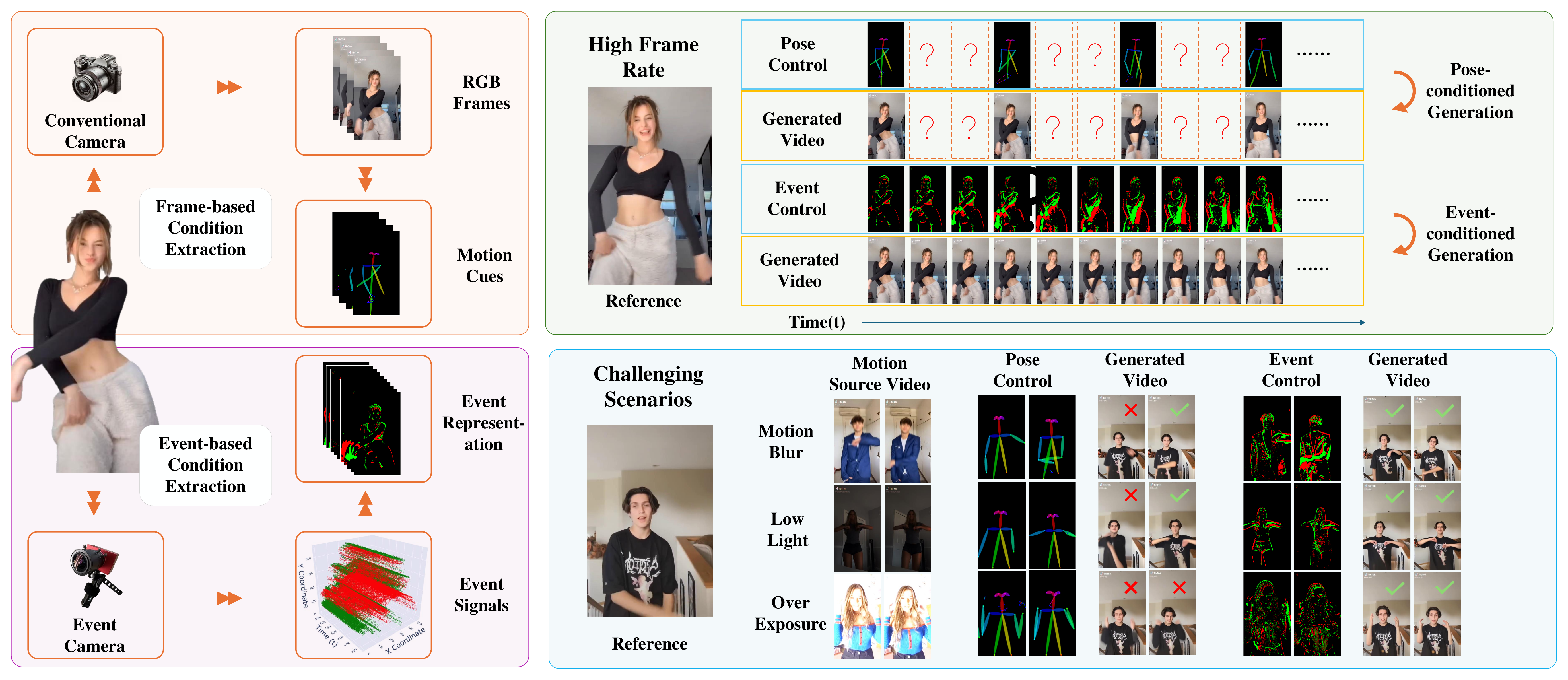} % Replace with your actual figure
    % \vspace{-1.8em}
    \captionof{figure}{\textbf{Comparison between the conventional image-to-video methods and the proposed EvAnimate framework.} \textbf{EvAnimate} leverages event streams as motion cues to generate controllable videos at high temporal resolutions. Moreover, \textbf{EvAnimate} produces superior video quality and exhibits enhanced robustness under challenging scenarios such as motion blur, low-light, and overexposure.}
    \label{fig:teaser}
\end{center}
}]

% In contrast to traditional approaches
% --- --- --- --- --- --- --- --- --- --- --- ---
%                      Abstract
% --- --- --- --- --- --- --- --- --- --- --- ---

% \input{sec/0_abstract}

% \vspace{5em}

% \footnote{*Corresponding author.}

\begin{abstract}

\vspace{-1em}

Conditional human animation traditionally animates static reference images using pose-based motion cues extracted from video data. However, these video-derived cues often suffer from low temporal resolution, motion blur, and unreliable performance under challenging lighting conditions. In contrast, event cameras inherently provide robust and high temporal-resolution motion information, offering resilience to motion blur, low-light environments, and exposure variations.
In this paper, we propose \textbf{EvAnimate}, the first method leveraging event streams as robust and precise motion cues for conditional human image animation. Our approach is fully compatible with diffusion-based generative models, enabled by encoding asynchronous event data into a specialized three-channel representation with adaptive slicing rates and densities. High-quality and temporally coherent animations are achieved through a dual-branch architecture explicitly designed to exploit event-driven dynamics, significantly enhancing performance under challenging real-world conditions. Enhanced cross-subject generalization is further achieved using specialized augmentation strategies. To facilitate future research, we establish a new benchmarking, including simulated event data for training and validation, and a real-world event dataset capturing human actions under normal and challenging scenarios. The experiment results demonstrate that \textbf{EvAnimate} achieves high temporal fidelity and robust performance in scenarios where traditional video-derived cues fall short.

\end{abstract}

% --- --- --- --- --- --- --- --- --- --- --- ---
%                    Introduction
% --- --- --- --- --- --- --- --- --- --- --- ---
\vspace{-1em}
\section{Introduction}
\label{sec:intro}

Image animation transforms static images into dynamic visual content, enhancing storytelling and interactivity across entertainment, virtual reality, digital art, and online retail applications~\cite{Han2018VITON, Wang2018CPVTON, Chan2019EverybodyDance, Kim2018DeepVideoPortraits}. Recent advancements, particularly diffusion-based generative models, have significantly improved the realism and controllability of pose-guided human animations from static images~\cite{Ho2020DDPM, Dhariwal2021DiffusionBeatsGANs, song2020denoising, Hu2024AnimateAnyone, Xu2024Magicanimate, Peng2024ControlNext, zhang2024mimicmotion, tumanyan2023plug, zhu2024champ}. These methods commonly extract motion cues from conventional videos to control the animation process.

However, traditional video-based motion extraction faces critical limitations, illustrated in Figure~\ref{fig:teaser}. Specifically, the low frame rates of video data inherently restrict temporal fidelity, while challenging conditions like high-speed motion, low lighting, or overexposure frequently cause motion blur and inaccuracies in extracted poses. These limitations significantly reduce the robustness and practical applicability of existing animation approaches.

To overcome these constraints, we highlight the potential of event cameras~\cite{Serrano2013DynamicVisionSensor, tedaldi2016feature, Gallego2020EventVisionSurvey, Zheng2023EventVisionSurvey} for human animation. Event cameras asynchronously capture pixel-level brightness changes triggered by motion, inherently emphasizing movement. This unique capability offers exceptionally high temporal resolution (around 1$\mu$s) and wide dynamic range (up to 140 dB), robustly capturing rapid actions and effectively handling extreme lighting conditions~\cite{Serrano2013DynamicVisionSensor, tedaldi2016feature, su2024motion, yu2024eventps, lin2025self}. Thus, event-based data provides precise, robust, and high-frame-rate motion cues, significantly enhancing animation quality and consistency, particularly in challenging real-world scenarios.

Inspired by these advantages, we propose \textbf{EvAnimate}, the first framework utilizing event-camera data as motion cues to animate static human images. By leveraging event streams, our approach substantially improves animation quality, temporal fidelity, and robustness compared to video-derived signals. Experiments using simulated event data with diverse motion patterns and a newly collected real-world event dataset validate the practical effectiveness and versatility of our method. Our contributions are summarized as follows:
\begin{itemize}

\item We propose \textbf{EvAnimate}, a novel diffusion-based framework that leverages event-camera data for robust human image animation. Our method significantly improves video quality and temporal consistency, especially under challenging real-world conditions.

\item Our framework is fully compatible with diffusion models, enabled by a novel \textit{event representation} that converts asynchronous event streams into three-channel slices with adjustable slicing rates and densities. A dedicated \textit{dual-branch architecture} explicitly captures event-driven motion dynamics, delivering animations with enhanced temporal fidelity. Specialized augmentation techniques further improve cross-person generalization.

\item We establish a comprehensive \textit{evaluation protocol and benchmark datasets} specifically designed for event-conditioned human animation, including simulated events for controlled training and a real-world dataset covering both typical and challenging scenarios. All datasets and code will be publicly available to facilitate future research.
\end{itemize}

% --- --- --- --- --- --- --- --- --- --- --- ---
%                   Related Work
% --- --- --- --- --- --- --- --- --- --- --- ---

\section{Related Work}
\label{sec:related work }

\subsection{Image Animation and Diffusion Models}
Image animation seeks to generate dynamic videos from static images under various controls (e.g., poses~\cite{Xu2024Magicanimate,Hu2024AnimateAnyone,Chang2023MagicPose}. Among these, pose control is the most prevalent, although direct pose-to-video methods often capture only instantaneous motion states~\cite{Ni2023LatentFlowDiffusion}. To address this limitation, dense motion prediction networks~\cite{Siarohin2019FirstOrderMotion,Zhang2023ConditionalControl} estimate dense heatmaps from sparse keypoints, and the Latent Flow Diffusion Model (LFDM)~\cite{Ni2023LatentFlowDiffusion} further leverages time-coherent flow in latent space for improved spatiotemporal consistency. Meanwhile, advances in diffusion models~\cite{SohlDickstein2015ThermodynamicsLearning,Dhariwal2021DiffusionBeatsGANs} have enabled high-fidelity image generation~\cite{Avrahami2022BlendedDiffusion,Rombach2022LatentDiffusion} with enhanced control via architectures like ControlNet~\cite{Zhang2023ConditionalControl} and T2I-Adapter~\cite{Mou2024T2iAdapter}. Extending diffusion to video, a 3D U-Net–based Video Diffusion Model (VDM)~\cite{Ho2022VideoDiffusion} captures spatiotemporal dynamics, while ControlNet-based methods~\cite{Chang2023MagicPose} further customize outputs. However, these approaches generally depend on reference videos or synthetic control signals, limiting their robustness in challenging settings. Stable Video Diffusion~\cite{Blattmann2023StableVideoDiffusion}, an adaptation of Stable Diffusion~\cite{Rombach2022LatentDiffusion} to 3D latent structures, addresses efficiency constraints for video generation. In this work, we adopt Stable Video Diffusion as our backbone.

\subsection{Event Camera}

The event camera, also known as a Dynamic Vision Sensor (DVS), has shown impressive performance in computer vision, especially in applications requiring high-speed motion detection or challenging lighting conditions. Unlike conventional cameras, each pixel in an event camera operates independently and triggers an event only when a brightness change is detected (caused by motion or flickering) \cite{Serrano2013DynamicVisionSensor, Zhu2018EVFlowNet,Gehrig2019AsynchronousLearning, qu2024evrepsl}. Specifically, an event is a tuple $(x_k, y_k, t_k, p_k)$, where coordinates $(x_k, y_k)$ identify the pixel that triggered the event, $t_k$ is the event timestamp, and $p_k \in \{1, -1\}$ represents the event polarity \cite{tedaldi2016feature, Gallego2020EventVisionSurvey, Chen2024Video2Haptics, su2024motion}. The sparse data captured by event cameras reveals only the outlines of moving objects, offering the advantage of low latency \cite{Zheng2023EventVisionSurvey, qu2024e2hqv, yu2024eventps, lin2025self}. Moreover, this characteristic allows event cameras to efficiently represent the essence of motion without requiring additional feature extraction \cite{Gallego2020EventVisionSurvey, Chen2024Video2Haptics, zhou2024evdig}, making them naturally suitable as control signals for image animation and pose transfer. In contrast, frame-based cameras record complete images or videos of a scene at fixed frame rates. Event cameras also offer other advantages, such as high temporal resolution (up to 1$\mu$s) and a high dynamic range (typically up to 140dB) \cite{Gallego2020EventVisionSurvey, barchid2023exploring, xu2024event, qu2024evrepsl}. This means that event cameras can capture data at ultra-high frame rates, even under challenging lighting conditions or during fast motion~\cite{Deng2020AMAE, cao2024embracing, qu2024e2hqv}, thereby significantly improving the robustness of motion extraction in scenarios where the quality of frames is affected. In summary, event cameras can provide precise, latency-free motion cues, making them an ideal complement to conventional frame-based methods for robust control signals in image animation and pose transfer.

% --- --- --- --- --- --- --- --- --- --- --- ---
%                   Methodology
% --- --- --- --- --- --- --- --- --- --- --- ---

\section{Methodology}

\begin{figure}[ht]
\centering{}
\includegraphics[width=1\linewidth]{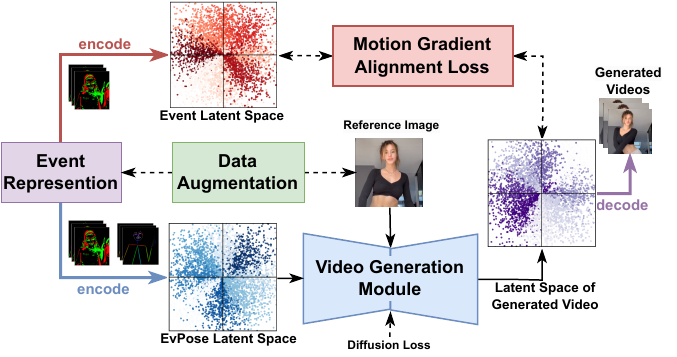}  
\caption{\textbf{Overview of the proposed event-conditioned human animation framework.}}   
\label{fig:overview}
\end{figure}

\subsection{Overview of EvAnimate}

In this section, we present \textbf{EvAnimate}, a framework that leverages event streams as motion cues within diffusion models to animate static human images (see Figure~\ref{fig:overview}). \textbf{EvAnimate} adopts a dual-branch architecture for enhanced consistency. First, the event streams are transformed into specially designed \textbf{Event Representations} that are suitable for diffusion models (detailed in Section~\ref{subsec:event_representation}). In the main branch (bottom flow in Figure~\ref{fig:overview}), event representations are fused with pose estimates derived from the events and then interactively encoded into an EvPose latent space. This latent space, combined with a reference image constraint, directs the \textbf{Video Generation Module} (described in Section~\ref{subsec:video_generation_module}) to create a video latent space trained using diffusion loss, which is subsequently decoded into actual videos. In the parallel branch (top flow in Figure~\ref{fig:overview}), we introduce an additional training cue, the \textbf{Motion Gradient Alignment Loss}, which leverages motion trends across sequential event slices to enforce motion consistency in the generated videos (see Section~\ref{subsec:mga_loss}). In addition, we design specialized \textbf{Data Augmentation} strategies to enhance cross-person generalization (detailed in Section~\ref{subsec:data_augmentation}), which is a critical step since the training dataset does not include event stream and reference image pairs from different persons. Importantly, by performing most operations within the latent space, \textbf{EvAnimate} significantly reduces the computational cost of the generation process.

\subsection{Event Representation}
\label{subsec:event_representation}

\textbf{Event cameras} operate asynchronously, responding to local changes in brightness and producing a continuous stream of events in the form \( (t, x, y, p) \)~\cite{Serrano2013DynamicVisionSensor, Gallego2020EventVisionSurvey, yu2024eventps}. The pixel location \((x,y)\), timestamp \(t\), and polarity \(p \in \{-1, 1\}\) jointly encode how and when the intensity at a given pixel crosses a preset threshold~\cite{Zheng2023EventVisionSurvey, qu2024e2hqv, zhou2024evdig, lin2025self}. Specifically, events arise if the logarithmic intensity \(L(t, x, y)\) changes by at least \(\pm C\) since the last event at \((x, y)\). To effectively utilize asynchronous event streams as conditioning inputs in diffusion models, it is essential to convert these streams into a suitable representation that meets \textit{three key requirements}:
\begin{itemize}
    \item \textit{Three-channel Format}: The representation should yield a three-channel slice that is compatible with diffusion models.
    \item \textit{Controllable Slicing Rate}: The number of event slices per second should be adjustable to match training frames.
    \item \textit{Appropriate Slice Density}: Each slice should include an appropriate number of events to produce a clear and meaningful image at every temporal resolution, ensuring that lower temporal resolutions are not overly dense and higher ones are not excessively sparse.
\end{itemize}

\begin{figure}[ht]
\centering{}
\includegraphics[width=1\linewidth]{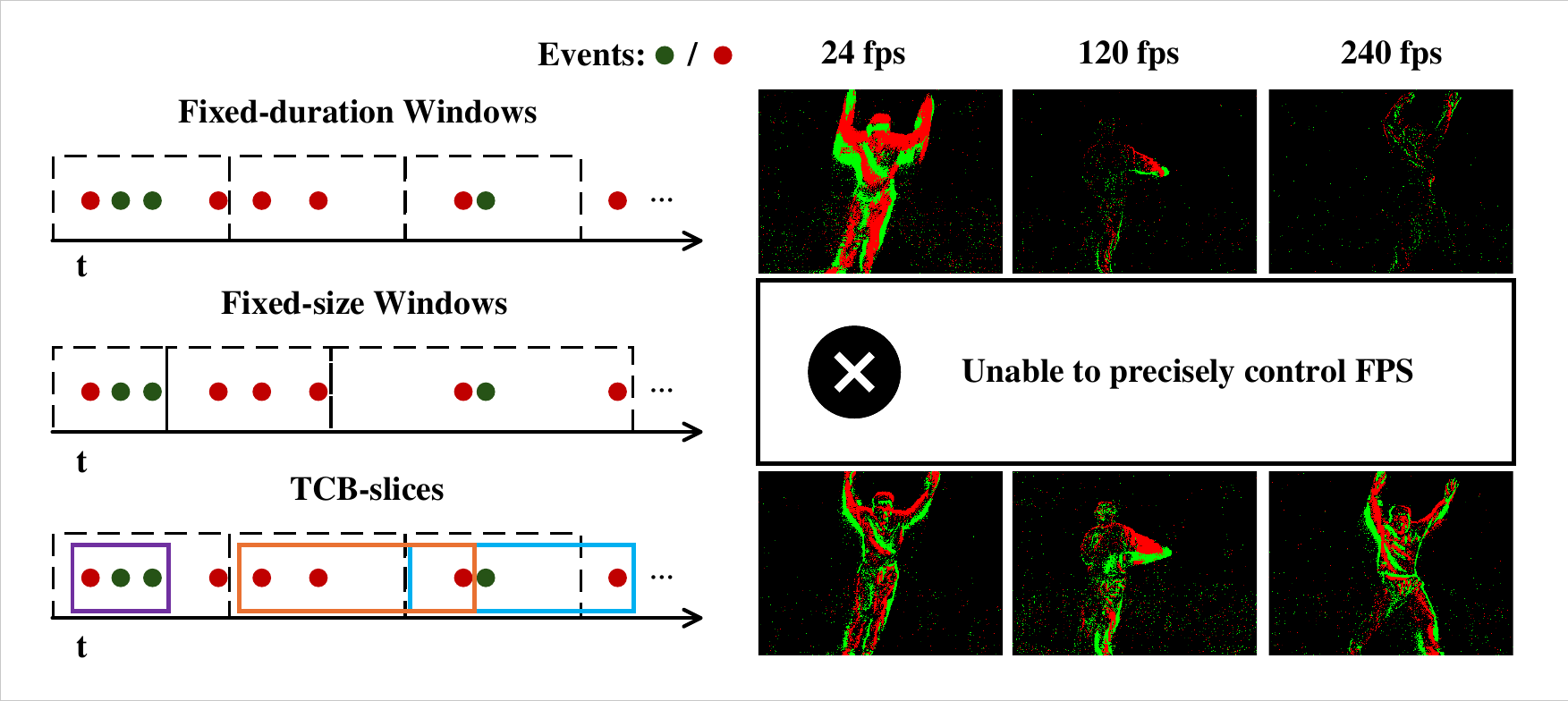}  
\caption{\textbf{Comparison of Event Representations.} The proposed TCB‐slices avoid the over-dense output seen with fixed-duration windows at low frame rates and the sparse output at high frame rates, while also enabling precise control over the frame rate compared to fixed-size windows.}   
\label{fig:tcb_slice}
\end{figure}

\paragraph{Time--Count Balanced slice (TCB-slice)} is designed to satisfy \textit{these requirements} (listed in the previous paragraph) by combining fixed-duration and fixed-size windows and balancing temporal resolution with event count, thereby mitigating blur at lower slice rates and sparsity at higher ones (as demonstrated in Figure~\ref{fig:tcb_slice}). Let \(F\) be the target slices per second, so that the base time interval is \(\Delta t = 1/F\). Define a threshold \(\Theta\) to distinguish low from high event slice rates. Let \(H\) and \(W\) be the height and width of the event camera, and let \(\alpha\) be a ratio that sets an event count per slice: \(M = \alpha H W\). For the \(i\)-th event slice, we start with the nominal time window \(\bigl[t_i,\, t_{i+1}\bigr)\), where \(t_{i+1} = t_i + \Delta t\), and collect all events with timestamps in this interval and $M$ events with timestamps after $t_i$. The final event set \(\Omega_i\) can be defined as:

\begin{equation}
\label{eq:event_window}
\begin{aligned}
\Omega_i \;= \; 
&\bigl\{ (t_k, x_k, y_k, p_k)\,\mid\, t_i \le t_k < t_{i+1} \bigr\} \;\otimes\; \\
&\bigl\{ (t_k, x_k, y_k, p_k)\,\mid\, t_i \le t_k,\;\text{until }|\Omega_i| = M \bigr\},
\end{aligned}
\end{equation}
where the assembling operator \(\otimes\) depends on the relation between target number of slices per second \(F\) and \(\Theta\) and is defined as:
\begin{equation}
\otimes =
\begin{cases}
\cap, & \text{if } F < \Theta, \\
\cup, & \text{if } F \ge \Theta.
\end{cases}
\end{equation}
This formulation ensures that the slice construction adapts to the target slicing rate by either constraining (via intersection) or augmenting (via union) the set of events, thus preserving the quality of the representation. In practice, we dynamically set $\Theta$ based on the median density of recent event slices for real-time adaptivity. This strategy effectively handles variations in event density. In our experiments, it typically ranges between 20 and 50. Finally, we accumulate the event polarities at each pixel \((x, y)\), denoted by \(\varphi_i\). Then, using these values, we create a three-channel slice \(\mathbf{C}_i\) by assigning colors according to the polarity sum: red for negative, green for positive, and black for zero. This TCB-slice formulation guarantees that each event slice is distinct while maintaining a controllable temporal slicing rate.

\subsection{Video Generation Module}
\label{subsec:video_generation_module}

\paragraph{Problem Definition:}
The aim of the EvAnimate is to generate event-conditioned high-quality video, preserving the content (e.g., characteristics of the human) from the reference image. Formally, let $\epsilon \sim N(0, I)$ denote a Gaussian noise volume with dimensions $K \times H \times W \times C$, where $K$, $H$, $W$, and $C$ represent the length, height, width, and channel number, respectively. Given an input image $x_0$ and event condition $e = \{e_1, e_2, \ldots, e_n\}$, the objective is to learn a mapping that transforms the noise volume $n$ into a sequence of frames $\{x_1, \ldots, x_K\}$, which are controlled by $x_0$ and $e$, ensuring that the distribution of the generated video matches that of the real video, i.e., $p(\hat{x}_{1:K} | x_0, e) \approx p(x_{1:K} | x_0, e)$. For the generation model, we employ a diffusion model~\cite{Dhariwal2021DiffusionBeatsGANs, Ho2022VideoDiffusion}. To reduce computational cost, we use latent diffusion~\cite{Rombach2022LatentDiffusion}: rather than applying the diffusion process directly on images, we first project them into a compact latent space using a variational autoencoder (VAE)~\cite{kingma2019introduction}, and then perform diffusion in that space. Formally, using a VAE structure, the encoder \(E\) maps each frame of a video sequence to a latent representation \(z = E(x)\), and the decoder \(D\) reconstructs the video sequence as \(x_{recon} = D(z)\). The primary objective of event-conditioned video generation is to learn an appropriate model \(\epsilon_\theta^*\) that minimizes the diffusion loss, i.e.:

\begin{equation}
\label{eq:obj}
\begin{aligned}
\epsilon_\theta^* = \argmin_{\theta}\frac{1}{N}\sum_{i=1}^{N} \Bigl\|\epsilon^i - \epsilon_\theta(z_t^i, e^i, t)\Bigr\|^2_2,
\end{aligned}
\end{equation}

\noindent where \(N\) is the total number of training samples, \(z_t^i\) is the noisy latent representation at timestep \(t\) (obtained via the forward diffusion process), \(e^i\) is the associated event condition (e.g., a set of event signals), and \(\epsilon^i \sim \mathcal{N}(0,I)\) is the injected Gaussian noise. The loss function (L2 norm) measures the discrepancy between the true noise \(\epsilon^i\) and the noise predicted by the model \(\epsilon_\theta(z_t^i, y^i, t)\). This formulation ensures that the reverse diffusion process, guided by both temporal and event conditions, generates video frames that are consistent with the input content and desired motion dynamics.

\begin{figure*}[ht]
\centering{}
\includegraphics[width=1\textwidth]{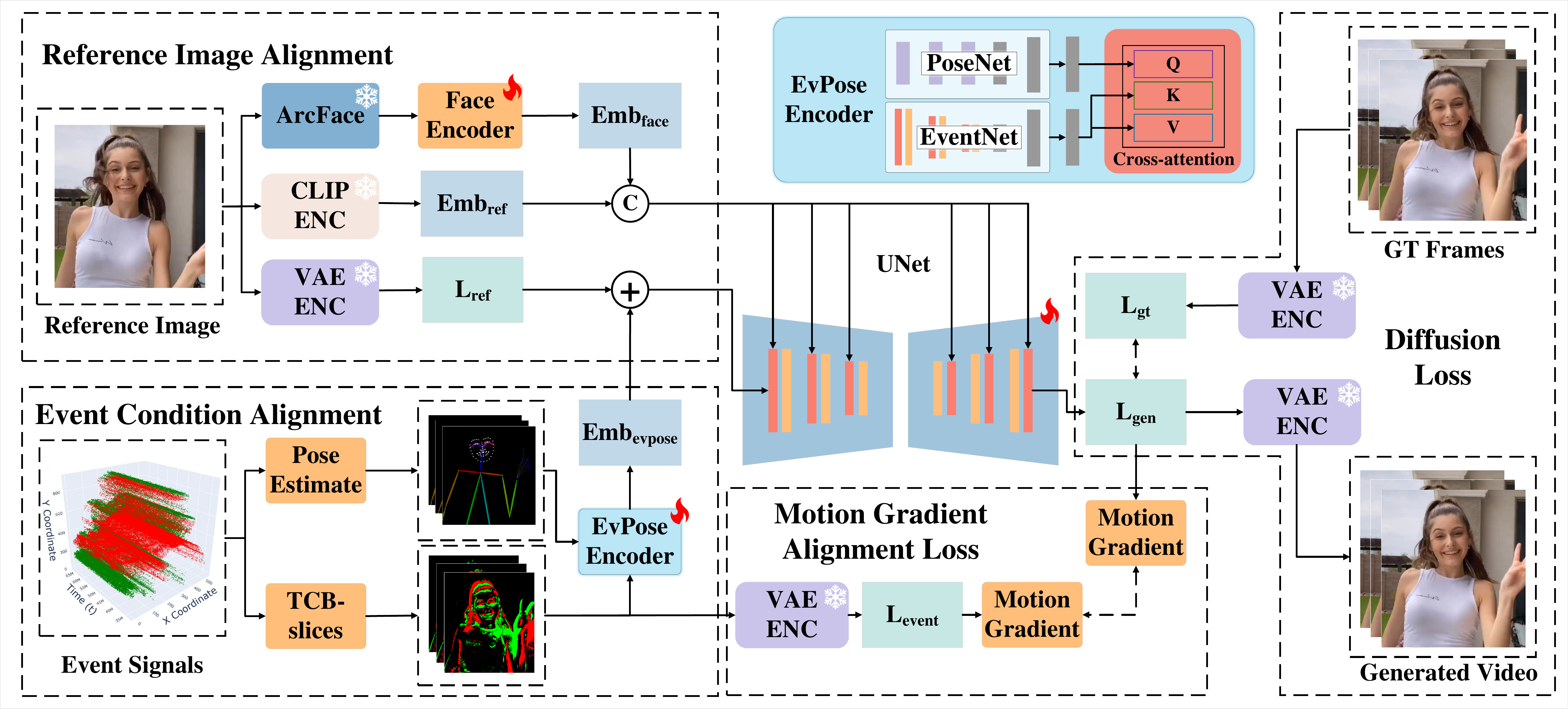}  
\caption{\textbf{Structure of the video generation module.} At its core, a spatial-temporal UNet generates latent representations of video frames. Four key components guide the process: (1) \textit{Reference Image Alignment} preserves the visual characteristics of the input by projecting the reference image into the latent space via a VAE and integrating semantic features from CLIP and face encoders; (2) \textit{Event Condition Alignment} controls motion by estimating pose from event signals and jointly encoding pose and event representations using a dual-encoder (EvPose Encoder); (3) \textit{Diffusion Loss} serves as the primary training objective by matching the latent representations of generated and ground truth videos; and (4) \textit{Motion Gradient Alignment Loss} leverages event conditions to enforce consistent, realistic motion dynamics.}   
\label{fig:framework}
\end{figure*}

\paragraph{Structure of the Video Generation Module:}  
Figure \ref{fig:framework} illustrates the overall architecture of the video generation module. At its core, the module employs a spatial-temporal UNet~\cite{Ronneberger2015UNet,Cicek2016ThreeDUNet} to generate latent representations of video frames. Four key components guide and control the generation process: \textit{Reference Image Alignment}, \textit{Event Condition Alignment}, \textit{Diffusion Loss}, and \textit{Motion Gradient Alignment Loss}. The \textit{Reference Image Alignment} component is designed to preserve the characteristics of the reference image. It uses a VAE encoder to project the reference image into the latent space, concatenates the resulting latent representation with diffusion noise, and supplies this combined input to the UNet. In addition, the reference image is processed by a CLIP encoder~\cite{radford2021learning} and a face encoder to extract semantic and facial features, which are then integrated into the UNet via cross-attention. The \textit{Event Condition Alignment} component is designed to control motions of the generated videos. It begins by estimating a pose from the event signals~\cite{yang2023effective}. This pose, along with event representations, is passed through a dual-encoder architecture (the EvPose Encoder) that employs cross-attention to jointly encode latent features for both pose and event data. The event representations are also fed into a VAE encoder to support the computation of the motion gradient alignment loss. The \textit{Diffusion Loss} measures the discrepancy between the latent representations of the generated videos and those of the ground truth videos, serving as the primary training objective. Finally, the \textit{Motion Gradient Alignment Loss} (detailed in Section~\ref{subsec:mga_loss}) is designed to fully leverage the event condition, ensuring consistent and realistic motion dynamics in the generated video.

\subsection{Motion Gradient Alignment Loss}
\label{subsec:mga_loss}

To fully leverage the event condition for consistent, aligned motion, we propose the Motion Gradient Alignment Loss (MGA-loss) alongside the diffusion loss in Equation~\ref{eq:obj}. The key idea is to align the generated video's motion dynamics with those inferred from event-conditioned latent representations. Rather than working in high-dimensional pixel space, MGA-loss processes compact latent features. It computes normalized temporal gradients via a center-difference kernel to capture fine-grained motion, then applies a contrastive formulation that emphasizes matching temporal segments (positive pairs) and penalizes mismatched ones (negative pairs). This reduces computational overhead while promoting realistic, coherent motion. Formally, let $\mathbf{L}_{\text{gen}}, \mathbf{L}_{\text{event}} \in \mathbb{R}^{B \times T \times C \times H \times W}$ denote the latent representations of the generated videos and event streams, respectively, where \(B\) is the batch size, \(T\) the number of time steps, \(C\) the number of channels, and \(H \times W\) the spatial dimensions. For a latent sequence \(\mathbf{L}\), the temporal gradient \(\mathbf{G}\) is computed via a center-difference kernel as:
\begin{equation}
G_{b,t,c,h,w} = \sum_{i=-1}^{1} w_{i+2}\, L_{b,t+i,c,h,w}, \quad t=2,\dots,T-1,
\end{equation}
with kernel weights \(\mathbf{w} = [w_1, w_2, w_3]\) (e.g., \([0.3,\, 0.4,\, 0.3]\)). The gradients are then normalized along the channel dimension $\tilde{\mathbf{G}} = {\mathbf{G}}/{\|\mathbf{G}\|_2}$.
Denote the normalized gradients for the generated and event latent spaces by \(\tilde{\mathbf{G}}^{\text{gen}}\) and \(\tilde{\mathbf{G}}^{\text{event}}\), respectively. A spatiotemporal similarity matrix is computed as:
\begin{equation}
S_{b,i,j,h,w} = \left\langle \tilde{\mathbf{G}}^{\text{gen}}_{b,i,:,h,w},\, \tilde{\mathbf{G}}^{\text{event}}_{b,j,:,h,w} \right\rangle,
\end{equation}
for \(i,j \in \{1,\ldots,T-2\}\), where the diagonal entries \(S_{b,i,i}\) correspond to the similarity between matching temporal segments (positive pairs), and the off-diagonal entries represent non-matching segments (negative pairs). For each batch element, the contrastive loss~\cite{chen2020simple} is defined as:
\begin{equation}
\mathcal{L}_{\text{MGA}} = -\frac{1}{B (T-2)} \sum_{b=1}^{B} \sum_{i=1}^{T-2} \log \frac{\exp\left(\frac{S_{b,i,i}}{\tau}\right)}{\sum_{j=1}^{T-2} \exp\left(\frac{S_{b,i,j}}{\tau}\right)},
\end{equation}

\noindent with \(\tau\) being a temperature hyperparameter. The MGA-loss, \(\mathcal{L}_{\text{MGA}}\), is integrated into the overall learning objective in Equation~\ref{eq:obj} with a loss weight $\lambda$. By formulating MGA-loss on latent representations, we further ensure that the generated videos exhibit consistent and realistic motion dynamics.

\subsection{Cross-person Data Augmentation}
\label{subsec:data_augmentation}
A challenge arises when the event and reference images represent different persons, while the training dataset contains video, event, and reference triplets from the same person. In such cases, the diffusion model tends to preserve the position and body shape from the event data rather than adapting to the body shape of the reference image. To address this overfitting issue, we introduce two data augmentation methods. The first method applies a random translation to the event representations by introducing small, random optical flow, thereby reducing the model's dependency on the specific body shape present in the event data. The second method randomly crops the reference image while ensuring that the face remains intact (using face detection~\cite{paraperas2024arc2face}), which diversifies the positional alignment between the reference and event inputs. Together, these augmentation strategies enhance the model's ability to generalize across different persons by mitigating overfitting to fixed positional and shape attributes.

% --- --- --- --- --- --- --- --- --- --- --- ---
%                   Experiment
% --- --- --- --- --- --- --- --- --- --- --- ---

\section{Experiments}
In this section, we conduct comprehensive experiments to evaluate the effectiveness, robustness, and generalization capabilities of our proposed \textbf{EvAnimate} framework. We first describe our training procedure utilizing simulated event data, specifically addressing the construction of our EvTikTok dataset. Subsequently, we provide thorough evaluations using a newly curated real-world dataset, EvHumanMotion, across diverse and challenging scenarios. Extensive quantitative and qualitative comparisons with state-of-the-art methods demonstrate the clear advantages of EvAnimate, supported by detailed ablation studies and analyses of critical aspects such as cross-person generalization and simulated-to-real domain gap.

\subsection{Training on Simulated Event Dataset}

\noindent \textbf{Construction of EvTikTok Dataset:}
Publicly available event datasets are limited for the task of human animation, and most existing datasets are designed for other tasks such as detection and tracking~\cite{Gallego2020EventVisionSurvey, cao2024embracing, xu2024event, lin2025self}. One common way to resolve the data shortage is to use a video-to-event simulator, like ESIM~\cite{Rebecq2018ESIM}, to convert video data into event streams. Although these simulators cannot fully replicate all the characteristics of real events, resulting in a gap between simulation and real-world application, they are still valuable for training purposes~\cite{stoffregen2020reducing}. Therefore, we created a simulated human animation dataset, termed EvTikTok, by using ESIM~\cite{Rebecq2018ESIM} to convert the TikTok dataset~\cite{Jafarian2021HumanDepth} into event streams. The TikTok dataset~\cite{Jafarian2021HumanDepth} is one of the most popular resources in the field of human animation, consisting of approximately 350 dance videos featuring a single person performing moderate TikTok dance moves (compilations for each month). Each video is 10-15 seconds long and recorded at 30 frames per second (fps), resulting in more than 100K images. We manually filtered severely degraded samples from our simulated training set to ensure data quality. We convert the events into TCB-slices (described in \ref{subsec:event_representation}) with 30 fps to match the frame rate of the RGB video. Our TCB-slice mitigates blur by adaptively balancing event density and temporal resolution.

\noindent \textbf{Training Configuration:} We ran 30K steps on a server equipped with 20 core Intel Xeon Platinum 8457C processor, 100 GB of RAM, and four NVIDIA RTX 3090 GPUs, using a learning rate of 5e-4.

\begin{table*}[htb]
\centering
\caption{\textbf{Quantitative comparisons of methods across various scenarios}, including low light, overexposure, motion blur, and normal. The best results in each metric are highlighted in \textbf{bold}, while the second-best are \underline{underlined}. $\uparrow$ denotes that a higher value is better, and $\downarrow$ indicates that a lower value is preferable.}
\resizebox{\textwidth}{!}{%
\label{tab:different_scenarios}

\begin{tabular}{lcccccccccccccccc}
\toprule
\multirow{2}{*}{\textbf{Method}} & \multicolumn{4}{c}{\textbf{Low Light}} & \multicolumn{4}{c}{\textbf{Overexposure}} & \multicolumn{4}{c}{\textbf{Motion Blur}} & \multicolumn{4}{c}{\textbf{Normal}} \\
\cmidrule(lr){2-5} \cmidrule(lr){6-9} \cmidrule(lr){10-13} \cmidrule(lr){14-17}
& \textbf{SSIM $\uparrow$} & \textbf{PSNR $\uparrow$} & \textbf{LPIPS $\downarrow$} & \textbf{FVD $\downarrow$} 
& \textbf{SSIM $\uparrow$} & \textbf{PSNR $\uparrow$} & \textbf{LPIPS $\downarrow$} & \textbf{FVD $\downarrow$} 
& \textbf{SSIM $\uparrow$} & \textbf{PSNR $\uparrow$} & \textbf{LPIPS $\downarrow$} & \textbf{FVD $\downarrow$} 
& \textbf{SSIM $\uparrow$} & \textbf{PSNR $\uparrow$} & \textbf{LPIPS $\downarrow$} & \textbf{FVD $\downarrow$} \\
\midrule
AnimateAnyone (CVPR24)~\cite{Hu2024AnimateAnyone} 
& 0.0245 & 6.0610 & 0.6811 & \underline{794.06} 
& \underline{0.6958} & \underline{13.8902} & \textbf{0.3131} & \underline{1208.32}
& 0.3169 & 7.6429 & 0.4917 & 1785.96
& 0.3163 & 7.7122 & 0.4863 & 1849.77 \\
Magicanimate (CVPR24)~\cite{Xu2024Magicanimate} 
& 0.0256 & 4.1862 & 0.8686 & 2642.16 
& 0.4063 & 7.7146 & 0.5414 & 2280.53
& 0.4919 & 13.4368 & 0.4055 & 1815.23
& 0.4661 & 12.9337 & 0.4397 & 1856.93 \\
MagicPose (ICML24)~\cite{Chang2023MagicPose} 
& \underline{0.1865} & 8.7518 & \underline{0.6040} & 2648.41 
& 0.5785 & 9.4112 & 0.4395 & 1562.29
& \underline{0.5952} & \underline{15.0975} & \underline{0.3743} & 1487.41
& \underline{0.6156} & \underline{14.7677} & \underline{0.3578} & 1514.33 \\
StableAnimator (CVPR25)~\cite{tu2024stableanimator} 
& 0.1399 & \underline{9.9106} & 0.6545 & 927.68 
& 0.3912 & 8.7813 & 0.5648 & 1581.96
& 0.4951 & 14.6244 & 0.3764 & \underline{745.53}
& 0.4831 & 14.1874 & 0.3888 & \underline{814.31} \\
\textbf{EvAnimate (Ours)} 
& \textbf{0.2925} & \textbf{23.3019} & \textbf{0.3211} & \textbf{429.81} 
& \textbf{0.7445} & \textbf{16.5443} & \underline{0.3201} & \textbf{650.45}
& \textbf{0.6959} & \textbf{19.5849} & \textbf{0.2469} & \textbf{691.37}
& \textbf{0.6843} & \textbf{19.1684} & \textbf{0.2565} & \textbf{638.51} \\
\hline
\textbf{Against the second-best} 
& \textbf{+56.8\%} & \textbf{+135.1\%} & \textbf{-46.8\%} & \textbf{-45.9\%}
& \textbf{+7.0\%} & \textbf{+19.1\%} & \textbf{+2.2\%} & \textbf{-46.2\%}
& \textbf{+16.9\%} & \textbf{+29.7\%} & \textbf{-34.0\%} & \textbf{-7.3\%}
& \textbf{+11.2\%} & \textbf{+29.8\%} & \textbf{-29.3\%} & \textbf{-21.6\%} \\
\bottomrule
\end{tabular}

}
\end{table*}

\subsection{Evaluating on Real-World Event Datasets}

\paragraph{Construction of EvHumanMotion Dataset:}

To rigorously evaluate human animation quality in real-world events, we have curated a comprehensive dataset of human actions using the DAVIS346 event camera~\cite{tedaldi2016feature}. This dataset integrates high-resolution RGB videos with precise event-based data, capturing both conventional and event-driven modalities. It comprises 113 sequences recorded from a balanced group of 20 participants (10 females and 10 males), ensuring a broad representation of human actions. The recordings span diverse settings, including indoor and outdoor environments as well as day and night conditions, and focus on dynamic movements, particularly dance sequences, to provide a challenging benchmark. Each sequence is approximately 10 seconds long at 24 fps, ensuring detailed temporal resolution. Furthermore, to facilitate a thorough evaluation of the proposed method under varied visual conditions, the dataset is systematically organized into four distinct scenarios: normal conditions (22 sequences), motion blur (22 sequences), overexposure (22 sequences), and low-light (47 sequences). Overall, this multifaceted and detailed dataset offers a robust resource for comprehensive evaluation in real-world settings.

\noindent \textbf{Evaluation Metrics:} We assess quality at both the video and frame levels. For video-level evaluation, we compute FID~\cite{Balaji2019ConditionalGAN}, IS~\cite{barratt2018note}, and FVD \cite{Unterthiner2018VideoMetrics} to quantify both spatial fidelity and temporal consistency. For frame-level evaluation, where one-to-one references are available, we report the average SSIM \cite{Wang2004SSIM}, PSNR, and LPIPS \cite{zhang2018unreasonable} as performance metrics. For SSIM \cite{Wang2004SSIM}, PSNR, and IS \cite{barratt2018note}, higher values indicate superior generation quality. Conversely, lower values for LPIPS \cite{zhang2018unreasonable}, FID \cite{Balaji2019ConditionalGAN}, and FVD \cite{Unterthiner2018VideoMetrics} denote better quality. All evaluations were conducted on the proposed EvHumanMotion dataset.

\subsection{Comparison with Other Human Animation Methods}
We evaluate the proposed \textbf{EvAnimate} against several state-of-the-art human animation approaches, including AnimateAnyone~\cite{Hu2024AnimateAnyone}, Magicanimate~\cite{Xu2024Magicanimate}, MagicPose~\cite{Chang2023MagicPose}, and StableAnimator~\cite{tu2024stableanimator} on the EvHumanMotion dataset. Our comprehensive comparison covers a wide range of scenarios from normal to challenging conditions, and examines both same-person and cross-person settings between event streams and reference images. Furthermore, we assess performance across different temporal resolutions.

\paragraph{Evaluation across Various Scenarios:}
We evaluate our method under four scenarios (i.e., low light, overexposure, motion blur, and normal), and summarize the quantitative results in Table~\ref{tab:different_scenarios}. As shown in Table~\ref{tab:different_scenarios}, our method consistently outperforms all competitors across most metrics, particularly excelling under challenging scenarios such as low light. Specifically, in the low light scenario, our method boosts SSIM by 56.8\% and PSNR by 135.1\% relative to the second-best approach. For overexposure, it demonstrates a significant 46.2\% reduction in FVD. In the motion blur scenario, our method achieves a 29.7\% increase in PSNR and a 34.0\% reduction in LPIPS. Even under the normal scenario, it maintains superior performance, with a 29.8\% increase in PSNR and a 29.3\% reduction in LPIPS compared to the second-best method. These results confirm the robustness and superior performance of our approach across diverse and challenging scenarios.

\paragraph{Evaluation on Same-person and Cross-person Setting:}

\begin{table}[htb]
\centering
\caption{\textbf{Quantitative comparisons of methods under same-person and cross-person settings.} In each metric, the best-performing value is highlighted in \textbf{bold} and the second-best is \underline{underlined}. An upward arrow ($\uparrow$) indicates that a higher value is preferable, whereas a downward arrow ($\downarrow$) indicates that a lower value is advantageous.}
\resizebox{\columnwidth}{!}{%
\label{tab:cross_person}

\begin{tabular}{lcccccc}
\toprule
\multirow{2}{*}{\textbf{Method}} & \multicolumn{3}{c}{\textbf{Same-person}} & \multicolumn{3}{c}{\textbf{Cross-person}} \\
\cmidrule(lr){2-4} \cmidrule(lr){5-7}
& \textbf{SSIM $\uparrow$} & \textbf{PSNR $\uparrow$} & \textbf{LPIPS $\downarrow$} 
& \textbf{FID $\downarrow$} & \textbf{IS $\uparrow$} & \textbf{FVD $\downarrow$} \\
\midrule
AnimateAnyone~\cite{Hu2024AnimateAnyone} & 0.2648 & 8.2689 & 0.5591 & 294.10 & 1.9182 & 2137.06 \\
Magicanimate~\cite{Xu2024Magicanimate} & 0.2625 & 8.5897 & 0.6135 & 267.23 & 1.7650 & 3444.27 \\
MagicPose~\cite{Chang2023MagicPose} & 0.3653 & 10.5567 & 0.5359 & 278.94 & 2.0911 & 3472.82 \\
StableAnimator~\cite{tu2024stableanimator} & \underline{0.4117} & \underline{12.9737} & \underline{0.4774} & \underline{265.66} & \underline{3.3391} & \textbf{1552.90} \\
\textbf{EvAnimate (Ours)} & \textbf{0.7603} & \textbf{20.6400} & \textbf{0.2873} & \textbf{206.56} & \textbf{4.8729} & \underline{1735.51} \\
\bottomrule
\end{tabular}%
}
\end{table}

We compare our method with others under both same-person and cross-person settings (i.e., whether the event and reference images capture the same person). Because there are no one-to-one ground truths for the cross-person setting, we rely on FID~\cite{Balaji2019ConditionalGAN}, IS~\cite{barratt2018note}, and FVD \cite{Unterthiner2018VideoMetrics} to evaluate generative quality. As shown in Table~\ref{tab:cross_person}, our approach surpasses all competitors on most metrics across both settings, with a notably large advantage in the same-person setting. Specifically, in the same-person setting, our method achieves an 84.7\% improvement in SSIM and a 59.1\% improvement in PSNR compared to the second-best approach.

\paragraph{Evaluation on High Temporal Resolutions:}
We evaluate our method at high temporal resolutions of 96 fps and 192 fps, with quantitative results summarized in Table~\ref{tab:high_fps}. As shown, our method consistently outperforms the competition by a substantial margin. Specifically, at 96 fps, our approach achieves an 84.3\% increase in PSNR and a 47.5\% reduction in LPIPS compared to the second-best. At 192 fps, it demonstrates a 78.7\% improvement in SSIM and an 81.5\% increase in PSNR.

\begin{table}[htb]
\centering
\caption{\textbf{Quantitative comparison of methods at 96 fps and 192 fps.} In each metric, the best-performing value is highlighted in \textbf{bold} and the second-best is \underline{underlined}.}
\resizebox{\columnwidth}{!}{%
\label{tab:high_fps}
\begin{tabular}{lcccccccc}
\toprule
\multirow{2}{*}{\textbf{Method}} & \multicolumn{4}{c}{\textbf{96 fps}} & \multicolumn{4}{c}{\textbf{192 fps}} \\
\cmidrule(lr){2-5} \cmidrule(lr){6-9}
& \textbf{SSIM $\uparrow$} & \textbf{PSNR $\uparrow$} & \textbf{LPIPS $\downarrow$} & \textbf{FVD $\downarrow$} 
& \textbf{SSIM $\uparrow$} & \textbf{PSNR $\uparrow$} & \textbf{LPIPS $\downarrow$} & \textbf{FVD $\downarrow$} \\
\midrule
AnimateAnyone~\cite{Hu2024AnimateAnyone} & 0.3778 & 8.0907 & 0.4802 & 2248.55 & \underline{0.3725} & 7.9833 & \underline{0.4817} & 1994.74 \\
Magicanimate~\cite{Xu2024Magicanimate} & 0.2204 & \underline{10.3381} & 0.5462 & \underline{1209.52} & 0.3714 & 7.9835 & 0.4818 & 1996.94 \\
MagicPose~\cite{Chang2023MagicPose} & 0.1865 & 8.7518 & 0.6040 & 2648.41 & 0.0998 & 9.7210 & 0.5989 & 1549.74 \\
StableAnimator~\cite{tu2024stableanimator} & \underline{0.5141} & 9.8748 & \underline{0.4498} & 2154.18 & 0.2204 & \underline{10.3381} & 0.5462 & \underline{1209.52} \\
\textbf{EvAnimate (Ours)} & \textbf{0.6984} & \textbf{19.0570} & \textbf{0.2363} & \textbf{814.02} & \textbf{0.6657} & \textbf{18.7596} & \textbf{0.2528} & \textbf{1153.84} \\
\hline
\textbf{v.s. the second-best} & \textbf{+35.8\%} & \textbf{+84.3\%} & \textbf{-47.5\%} & \textbf{-32.7\%} & \textbf{+78.7\%} & \textbf{+81.5\%} & \textbf{-47.5\%} & \textbf{-4.6\%} \\
\bottomrule
\end{tabular}%
}
\end{table}

\begin{figure*}[htb]
\centering{}
\includegraphics[width=1\textwidth]{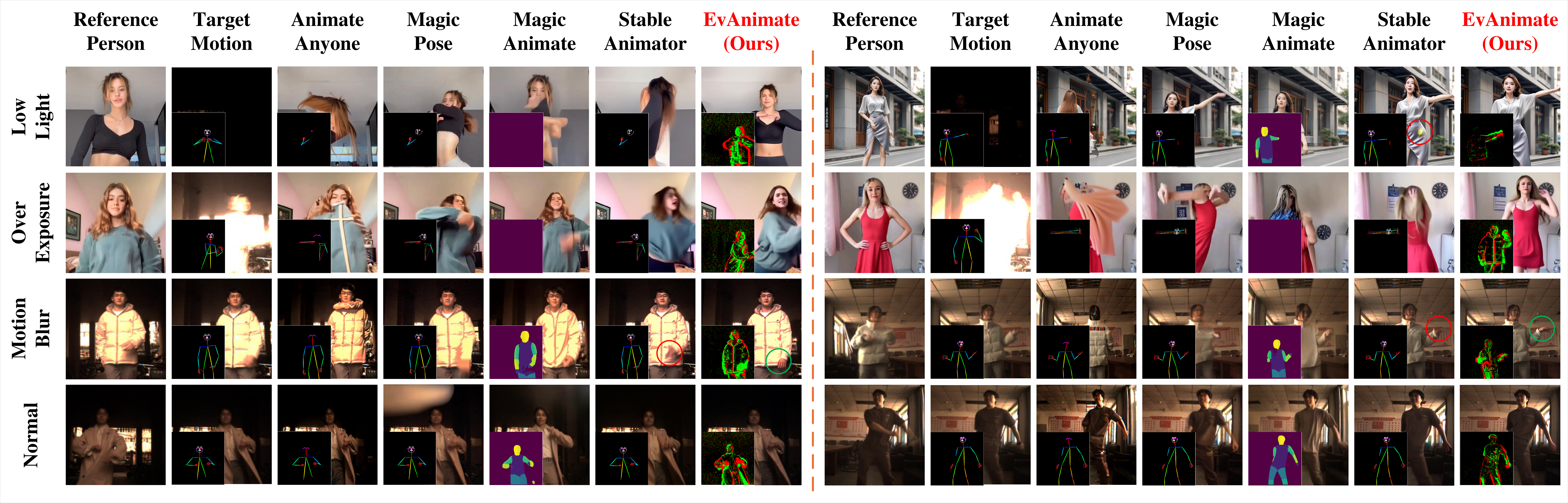}  

\caption{\textbf{Qualitative comparison of EvAnimate with state-of-the-art methods across various scenarios} (low light, overexposure, motion blur, normal). The first column shows the reference image, followed by animations generated by AnimateAnyone, MagicPose, MagicAnimate, StableAnimator, and our method. EvAnimate consistently achieves superior visual fidelity and accurate motion reproduction, especially under challenging conditions.}   
\label{fig:qualitative}

\end{figure*}

\begin{figure*}[htb]
\centering{}
\includegraphics[width=1\textwidth]{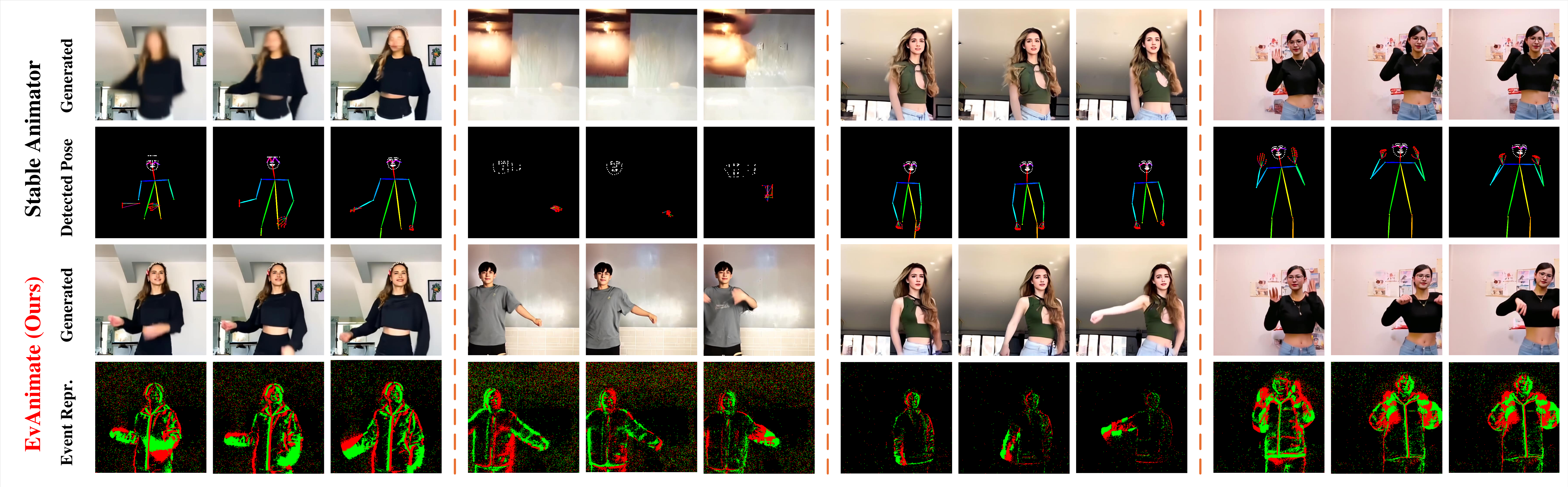}  

\caption{\textbf{Representative frame sequences demonstrating dynamic quality and temporal coherence.} Compared with the StableAnimator (i.e., the closest competing approach), EvAnimate produces significantly fewer artifacts, better preserves detailed motions (e.g., hand positions), and exhibits improved visual stability. Visualizations of detected poses and corresponding event representations highlight the effectiveness of our event-driven animation approach.}   
\label{fig:video}

\end{figure*}

% \vspace{-1em}

\subsection{Qualitative Results}

\paragraph{Cross-Scenario Quality:}
We present qualitative comparisons in Figure~\ref{fig:qualitative}, systematically covering diverse scenarios including low light, overexposure, motion blur, and normal lighting conditions. Each row depicts a specific scene with the first column showing the reference image and subsequent columns illustrating results from various state-of-the-art methods. Improved pose visualizations clarify the provided motion cues, highlighting critical details such as poses and hand movements. Notably, our proposed \textbf{EvAnimate} consistently preserves the subject’s appearance and generates motion sequences closely aligned with target motion cues, significantly outperforming competing methods especially under challenging conditions. These results clearly showcase the advantages of leveraging event data for human animation.

\paragraph{Dynamic Quality:}
To illustrate the temporal quality and coherence achievable by our method, representative frame sequences are provided in Figure~\ref{fig:video}. Compared to StableAnimator (i.e., the closest competing approach), which represents the closest competing approach, animations produced by \textbf{EvAnimate} exhibit notably reduced temporal artifacts, improved pose reproduction accuracy, and enhanced visual stability across consecutive frames. Furthermore, visualizations of event representations confirm that event data effectively captures robust and precise motion dynamics, underscoring the advantages of event-driven animation.

\subsection{Ablation Studies} 

\begin{table}[ht]

\centering
\caption{\textbf{Ablation study of different modules.} The best values for each metric are in \textbf{bold}.}
\resizebox{\columnwidth}{!}{%
\label{tab:ablation}

\begin{tabular}{lcccc}
\toprule
\textbf{Method} & \textbf{SSIM $\uparrow$} & \textbf{PSNR $\uparrow$} & \textbf{LPIPS $\downarrow$} & \textbf{FVD $\downarrow$} \\
\midrule
Backbone + Pose & 0.1187 & 5.8591 & 0.8245 & 1549.82 \\
Backbone + Event  & 0.1460 & 8.0616 & 0.7603 & 1335.63 \\
Backbone + Event + MGA-loss & \textbf{0.7603} & \textbf{20.6400} & \textbf{0.2873} & \textbf{516.44} \\
\bottomrule
\end{tabular}%
}
\end{table}

To assess the impact of our design choices in EvAnimate, we conduct ablation studies on the proposed modules, as shown in Table~\ref{tab:ablation}. We first compare the backbone’s performance when using poses versus events as motion cues. As shown in Table~\ref{tab:ablation}, the event-based approach performs better than the pose-based one. Moreover, Table~\ref{tab:ablation} shows that adding the proposed MGA-loss (introduced in Section~\ref{subsec:mga_loss}) provides a significant performance boost. We also examine the effectiveness of our cross-person data augmentation (introduced in Section~\ref{subsec:data_augmentation}). As shown in Table~\ref{tab:data_aug}, this data augmentation further enhances performance in cross-person settings.

\begin{table}[htb]
\centering
\caption{\textbf{Evaluation on effectiveness of cross-person data augmentation.} The best values are highlighted in \textbf{bold}}
\resizebox{\columnwidth}{!}{%
\label{tab:data_aug}

\begin{tabular}{lccc}
\toprule
\textbf{Method} & \textbf{FID $\downarrow$} & \textbf{IS $\uparrow$} & \textbf{FVD $\downarrow$} \\
\midrule
EvAnimate without Cross-person Data Aug. & 318.93 & 1.1205 & 2058.67 \\
EvAnimate with Cross-person Data Aug. & \textbf{206.56} & \textbf{4.8729} & \textbf{1735.51} \\
\bottomrule
\end{tabular}%
}
\end{table}
% \vspace{-1em}

\subsection{Simulated-to-Real Domain Gap Analysis}

Training on simulated event data and evaluating on real-world data is a common and deliberate strategy in event-camera research to prevent model overfitting to specific camera sensors or environmental conditions. Given the current lack of extensive real-world datasets tailored for event-based human animation, our training primarily relies on simulated data. Nevertheless, assessing the performance gap between simulated and real domains is crucial. We quantitatively examine this gap in Table~\ref{tab:domain_gap}, comparing performance on the simulated dataset (EvTikTok) and the real-world dataset (EvHumanMotion). The modest performance reduction from simulated to real-world settings demonstrates that our method effectively generalizes across domains, with only a limited domain gap. These results confirm the robustness of our approach and justify using simulated event data for training, particularly given the current scarcity of real-world event-based human animation data. Future work could further narrow this domain gap through targeted domain adaptation techniques or expanded real-world dataset collection.

\begin{table}[htb]
\centering
\caption{\textbf{Quantitative evaluation of the simulated-to-real domain gap.} The modest performance difference between simulated (EvTikTok) and real-world (EvHumanMotion) datasets indicates that EvAnimate effectively generalizes to real event data.}
\resizebox{\columnwidth}{!}{%
\label{tab:domain_gap}
\begin{tabular}{lccc}
\toprule
\textbf{Dataset} & \textbf{SSIM $\uparrow$} & \textbf{PSNR $\uparrow$} & \textbf{LPIPS $\downarrow$} \\
\midrule
Real-world (EvHumanMotion) & 0.6843 & 19.1684 & 0.2565 \\
Simulated (EvTikTok) & 0.724 & 20.3041 & 0.2012 \\
\bottomrule
\end{tabular}%
}
\end{table}

\section{Conclusion and Future Direction}

We presented \textbf{EvAnimate}, the first diffusion-based framework leveraging event-camera data to animate static human images. Our approach robustly transfers realistic motions from captured event signals to new target identities, a capability not achievable by direct video capture alone. By encoding event streams into three-channel representations and employing a dual-branch architecture, our method achieves superior animation quality and temporal consistency, particularly under challenging conditions such as motion blur, low lighting, and overexposure. Extensive evaluations on newly established real-world benchmarks demonstrate that \textbf{EvAnimate} significantly outperforms conventional methods, highlighting its practical applicability in uncontrolled environments like concerts, outdoor performances, and rapid-motion scenarios. While our method currently faces limitations due to biases inherent in the training data (mostly attractive females from TikTok), addressing these and incorporating interactive editing represent valuable directions for future research. To facilitate further progress, all datasets and code developed in this work will be publicly released.

{
    \small
    \bibliographystyle{ieeenat_fullname}
    \bibliography{main}
}

\end{document}